\pdfoutput=1
\documentclass[11pt]{article}

\usepackage{acl}

\usepackage{times}
\usepackage{helvet}  % DO NOT CHANGE THIS
\usepackage{courier}  % DO NOT CHANGE THIS
\usepackage{booktabs}

\usepackage{newfloat}
\usepackage{listings}
\usepackage{graphicx}

\usepackage[utf8]{inputenc}

\usepackage{microtype}

\usepackage{inconsolata}

\title{LLM-Based Section Identifiers Excel on Open Source but \\ Stumble in Real World Applications}

\author{Saranya Krishnamoorthy, Ayush Singh, Shabnam Tafreshi \\
  inQbator AI at eviCore Healthcare \\
  Evernorth Health Services\\
  \texttt{firstname.lastname@evicore.com}}

\begin{document}
\maketitle
\begin{abstract}
Electronic health records (EHR) even though a boon for healthcare practitioners, are growing convoluted and longer every day. Sifting around these lengthy EHRs is taxing and becomes a cumbersome part of physician-patient interaction. Several approaches have been proposed to help alleviate this prevalent issue either via summarization or sectioning, however, only a few approaches have truly been helpful in the past. With the rise of automated methods, machine learning (ML) has shown promise in solving the task of identifying relevant sections in EHR. However, most ML methods rely on labeled data which is difficult to get in healthcare. Large language models (LLMs) on the other hand, have performed impressive feats in natural language processing (NLP), that too in a zero-shot manner, i.e. without any labeled data. To that end, we propose using LLMs to identify relevant section headers. We find that GPT-4 can effectively solve the task on both zero and few-shot settings as well as segment dramatically better than state-of-the-art methods. Additionally, we also annotate a much harder real world dataset and find that GPT-4 struggles to perform well, alluding to further research and harder benchmarks. 
% However, their effectiveness at processing longer documents remains a bottleneck due to their inherent architecture. One emerging method to process long documents is to only fetch relevant parts of a document, given a query, via semantic search. The current search architecture to enable this has a lot of moving parts, is slow, and has a large preprocessing overhead.
\end{abstract}

\section{Introduction}

\begin{figure}[t]
    \centering
    \includegraphics[width=0.50\textwidth]{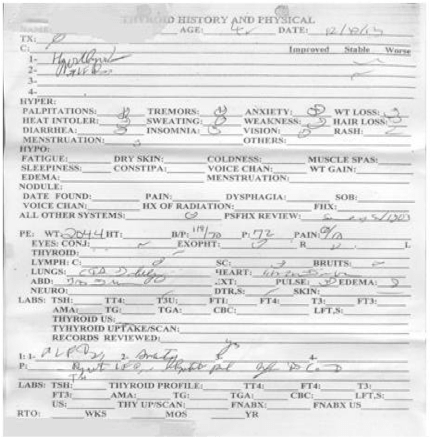}
    \caption{Sample real world obscure image of an outpatient paper-based patient encounter form comprising of numerous sections \cite{hersh2018health}.}
    \label{fig:real-world}
\end{figure} 

Modern day healthcare systems are increasingly moving towards large scale adoption of maintaining electronic health records (EHR) of patients \cite{congress2009hr}. EHRs help healthcare practitioners with relevant information about a patient such as history, medications, etc. However, in recent times this practice has led to very long and convoluted EHRs \cite{adam_lengthy_notes2021}. Naturally, the need for better information retrieval tools emerged due to the progressively lengthy and unstructured doctor notes. One such need is the accurate identification of sections in an EHR, pertinent to a physician's inquiry. For instance, a question like ``What treatments has the patient undergone in the past?'' concerning prior treatments administered to a patient necessitates the swift extraction of information from the ``treatments'' and ``past medical history'' sections, while excluding sections related to ``ancestral medical history''. This swift extraction is vital for timely decision-making in patient care. Additionally, during critical procedures such as the evaluation of medical necessity for prior authorization requests, it is customary for experienced clinicians to locate vital data within specific sections. An illustrative case entails examining the ``physical exam'' section to identify particular findings, such as signs of neurological disorders or movement-associated pain, indicating the need for additional diagnostic tests. The timely identification of such information is of utmost importance in ensuring the provision of appropriate care and reducing the risk of potential complications.

In general, regions found in EHR would often have a section heading preceding the body of the section, as can be seen in example Table \ref{tab:sample_id}. Even though these section types have limited cardinality, however, more often than not, physicians would fail to adhere to standards and use lexical variations generated on the fly. Moreover, practitioners not only will generate lexical variations of sections on the fly but also completely new sections altogether for valid reasons like imaging reports, etc. Apart from these variations, oftentimes there would be no headers at all, even though the information present could ideally be part of a pre-existing section in a document or a new section altogether. While studies like \citet{gao2022hierarchical} utilize the Subjective, Objective, Assessment and Plan heading (SOAP) framework, real-world clinical notes often contain sections beyond these categories. This limitation is further emphasized in \citet{landes2022new}, warranting further investigation and analysis.

The aforementioned factors have consequently contributed to the establishment of Section Identification (SI) as a distinct and enduring problem within the academic discourse \citep{McKnightS03}, making it an indispensable component of any clinical natural language processing (NLP) pipeline. A SI task entails finding regions of text that are semantically related to an aspect of a patient's medical profile. More importantly, it helps to improve pre-existing information retrieval systems by enabling them to be more targeted and specific. Lastly, in light of recent findings of the negative impact of note bloat within EHRs on even the most sophisticated systems \cite{LIU2022104149}, using SI to shorten or create from EHR, a sub-EHR specific to a given task would prove to be a worthwhile effort for humans and machines both. 

\begin{table*}[ht!]
    \centering
    \small
    \begin{tabular}{p{2cm}p{13cm}}
        \textbf{Allergies} &  {Allergies:} Patient recorded as having No Known Allergies to Drugs... \\ \midrule
         \textbf{History of Present Illness
} & {HPI}: 61M w/ incidental L renal mass found during W/U for brachytherapy for low-grade [**Last Name (STitle) **], now w/ gradually worsening gross hematuria for the past several days.\\ \midrule
       \textbf{Labs Imaging} & {Pertinent Results:} [**2160-4-10**] 07:30AM BLOOD WBC-12.6* RBC-3.20* Hgb-8.2* Hct-24.5*
MCV-77* MCH-25.6* MCHC-33.4 RDW-17.1* Plt Ct-438. \\ \midrule
        \textbf{Hospital Course}  & {Brief Hospital Course:}
61M w/ low-grade [**Month/Day/Year **] awaiting brachytherapy and locally-advanced L renal mass w/ collecting system invasion, renal vein thrombus, and likely metastases, presented w/gradually worsening gross hematuria.\\ \midrule
\end{tabular}
    \caption{ This figure illustrates a sample data point from the MIMIC-III database, highlighting the sections annotated with  MedSecID corpus. }
    \label{tab:sample_id}
\end{table*}

Because finding sections and hence their corresponding headers involves inherent variability, machine learning (ML) methods have played an important role in this natural language processing \cite{PomaresQuimbaya2019CurrentAT}. ML has increasingly been shown to be efficient in finding relevant sections within a document, however, a key drawback of traditional ML methods has been the dependence on labeled data \cite{tepper2012statistical}. Reliance on annotated data for training ML models to be able to predict the beginning and end of section headers has stalled the field from fully solving the task. The emergence of large language models (LLMs) in contemporary research presents a promising avenue to overcome the limitations inherent in traditional machine learning approaches, thereby expanding the scope of their applications.

LLMs have emerged as the de-facto system for NLP in scenarios where data is scarce \cite{openai2023gpt4}. The key distinction between traditional Machine Learning (ML) models and Large Language Models (LLMs) lies in their ability to understand tasks in natural language. While traditional ML models require labeled data for training, LLMs can leverage pre-training on vast amounts of unstructured text data, enabling them to perform tasks with minimal task-specific fine-tuning. This makes ML possible in an unsupervised manner (no need for labeled data) and therefore opens room for applications in domains where annotated data is hard to acquire like healthcare.
While LLMs have been evaluated on a wide array of NLP tasks in healthcare \cite{nori2023capabilities}, they are yet to be evaluated on their effectiveness in segmenting a document into semantically relevant sections.

In this work, we address this gap and evaluate the efficacy of our approach on a widely-known datasets in the clinical medical domain. Findings show that GPT-4 \cite{openai2023gpt4} almost solved the section identification problem on the benchmark open-sourced dataset, however, on a private dataset the performance lags. Our contributions are three-fold, listed as follows:  
% Second, because existing open source datasets are quite clean which is contrary to real world data which is messy, we annotate real-world data and evaluate GPT on it. Our findings show contrasting performance of GPT on open source compared to real-world datasets.

 \begin{enumerate}
     \item We show that GPT-4 can generate zero-shot headings of records with very high accuracy.
     \item Contrary to the above, we find that its performance drops on internal real-world datasets.
     \item An ontology of numerous section headers seen in real world EHR systems is shared which has much higher coverage.
     % \item GPT can generate zero-shot headings of records with very high accuracy
    % % \item 
  %  \item A dataset 10x larger than the current dataset of MedSecId generated using distant supervision
 \end{enumerate}

%LLMs are zero-shot section identifiers → we break the SoTA on MedSecId by using GPT.
\section{Related Work}
Traditionally, SI task has been done using a pre-defined dictionary of plausible candidates. \citet{PomaresQuimbaya2019CurrentAT} performed a comprehensive survey and found that rule-based methods still dominated the array of methods proposed while ML systems increasingly achieved better coverage when combined in a hybrid manner with rule-based methods. \citet{McKnightS03} later on extracted bag-of-words from MedLINE abstracts and used a support vector machine to train a classifier to categorize sentences into either Introduction, Method, Result, or Conclusion, demonstrating promising results. Similarly, \citet{hirohata-etal-2008-identifying} achieved very high accuracy by using conditional random fields to label scientific abstracts into Objectives, Methods, Results, and Conclusions.

Over time and with the inclusion of ML, the field re-framed this problem as one of span-level entity identification i.e. the system would be tasked with predicting whether each token in a sequence belongs to one of the predefined section types using the Inside-Outside-Beginning (IOB) tagging system \cite{ramshaw1999text}. \citet{tepper2012statistical} addresses the task of segmenting clinical records into distinct sections using a two-step approach. First, the section boundaries are identified. Then, the sections are passed to the second step, where a classifier is used to label each token as \textit{Begin}, \textit{In} or \textit{Out} of the span of a section.
\citet{nair2021clinical} proposes several transfer learning models based on clinical contextual embeddings for classifying clinical notes into the major SOAP sections \cite{soap-notes}.
\citet{zhou2023improving} investigates the effectiveness of continued pre-training in enhancing the transferability of clinical note section classification models. Both of the above papers resemble our work, however, they restrict them to SOAP sections and train specific models to do so. While the techniques devised so far have shown promise, to the best of our knowledge none of the previous works have tried in an unsupervised manner.

With the advent of LLMs \citep{devlin2018bert, openai2023gpt4}, several works have shown the efficacy of LLMs in doing unsupervised zero-shot information extraction. The primary method for interacting with generative LLMs is by the use of natural language prompts. \citet{wei_chain--thought_2022} found a significant performance boost by asking the model to explain its chain of thought before answering the query. Further, \citet{brown2020language} showed that additional performance can be gained by passing some examples as part of the prompt, they named it Few-Shot prompting. \citet{wang2023gpt, bian2023inspire, ashok2023promptner} have shown the efficacy of prompting the LLM to extract biomedical named entities from scientific articles. More recently, \citet{liu2023deidgpt} used GPT-4 to de-identify documents in a zero-shot manner. This hints at the immense document understanding capabilities of LLMs and opens doors to its application to a wide array of previously unresolved tasks such as SI. 

Apart from the advancements in the field of ML and SI, to evaluate how well SI systems perform, a standardization of tasks as well as datasets is required. To that end, \citet{uzuner20112010} first proposed a SI task as part of Informatics for Integrating Biology and the Bedside (i2b2) benchmarks. Recently, \citet{landes2022new} argued that the previous dataset did not fully cover the nuances in SI task and proposed a dataset an order of magnitude larger as well as more comprehensive than one by \citet{uzuner20112010}. However, the dataset proposed by \citet{landes2022new} is based on a clean source \citet{johnson2016mimic}, which oftentimes is not the case in real-world scenarios. To that end, we also annotated a real-world dataset to evaluate LLMs on it as well.

\section{Datasets}

\begin{table*}[ht]
    \centering
    \begin{tabular}{lccc}
    \textbf{Dataset} &  \textbf{MedSedId} & \textbf{i2b2 2010} & \textbf{Real World}   \\ \midrule
    Document count                 & 2002 & 96 &100   \\
    %Section labels & 50 &    \\
    Average token length           & 2307 & 1283 &7841 \\
    Std. dev. token length           & 1732 & 726 &8093 \\
    Average sections per doc & 12& 17  & 12  \\
    Std. dev. sections per doc    & 5.7  & 6.2 & 8 \\
    \end{tabular}
    \caption{Corpus Statistics}
    \label{tab:processbank}
\end{table*}

\subsection{i2b2 2010} 
In their study, \citet{tepper2012statistical} meticulously curated a corpus comprising 183 annotated clinical notes extracted from a selection of discharge summaries within the i2b2 2010 \cite{uzuner20112010} dataset. This dataset was annotated by an expert and served as a valuable resource for their research. However, owing to constraints imposed by Institutional Review Boards (IRBs), our current access to the i2b2 2010 dataset is limited. As a result, we were only able to procure clinical notes for 96 out of the originally annotated 183 documents.

\subsection{MedSecID}
MedSecID \citep{landes2022new} is a publicly available corpus of 2,002 fully annotated medical notes from the MIMIC-III \citep{johnson2016mimic} clinical record database. Each note has been manually annotated with section boundaries and section labels (See Table \ref{tab:sample_id} for an example of a typical clinical note consisting of well-defined sections). The section labels correspond to different types of information that are typically found in clinical notes, such as history of present illness, physical exam findings, and progress notes. 

\subsection{Real-world}
In an increasingly digital world, one would be inclined to assume healthcare data also lives digitally. Surprisingly, that is not the case almost 75\% of the healthcare dataset still lives in faxes \cite{faxburden} (see figure \ref{fig:real-world} for a sample handwritten and faxed clinical notes). Whereas all preexisting SI datasets are digitally derived from clean EHR systems, which even though offer us some insight into the performance of state of art, however, fail to paint the full picture. Therefore, we use an internal dataset of prior authorization requests derived from faxed-in images being transcribed to text via an optical character recognition system (OCR). These requests contain EHR of patients in the form of doctors' notes, submitted in both PDF and image formats. These documents lack a standardized structure, with segments and titles that can vary significantly in length. Although it's possible to group these titles into clusters of similar meaning, the language and number of titles differ across documents. Additionally, OCR inaccuracies arise from unclear text, spelling errors, complex table structures, and handwritten content, resulting in highly noisy input for any SI system to process.

\section{Annotation Methods}
\label{data-ann}

\begin{figure*}[ht!]
    \centering
    \includegraphics[width=\textwidth]{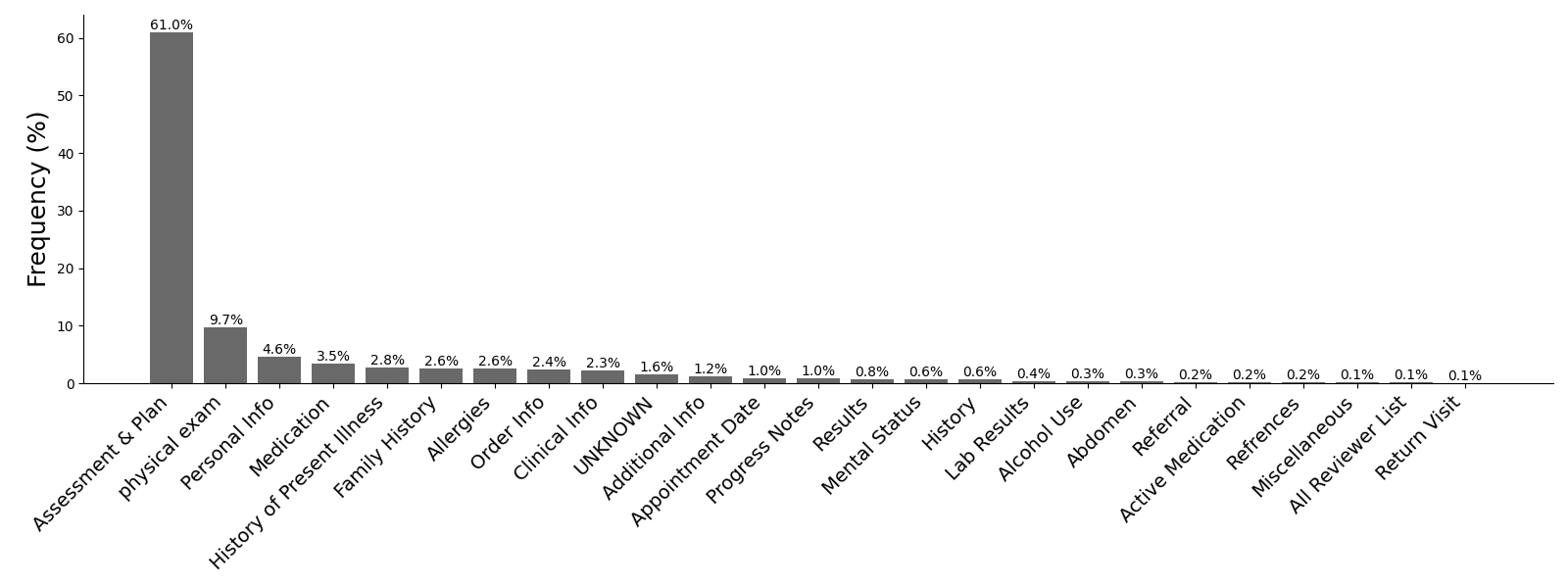}
    
    \caption{Section categories which are selected based on observation of top-header sections in the corpus and human judgment to associate section names to their topic or category of representations.}
    \label{fig:section}
\end{figure*} 

\begin{table*}[h]
    \centering
    \small
        \begin{tabular}{p{1.5cm}p{13cm}}\toprule
         \textbf{Medications Section} & Information about the current and past Medications\\ \midrule
         \textbf{Order Info} & This section consists of additional items that are required to conclude the assessments. Examples of such items are Mammograms, x-rays, etc., or the information about the provider of such items.\\ \midrule
         \textbf{Results Section} & Usually contains of lab results\\ \midrule
         \textbf{Physical Exam Section} & Result of physical exams such as Integumentary, Chest and Lung Exam, Cardiovascular, Abdomen, etc.\\ 
\end{tabular}
\medskip
    \caption{A sample of sections and subsections with the highest frequency.}
    \label{tab:sample_fre_cat}
    \begin{tabular}{p{1.5cm}p{13cm}}\toprule
         \textbf{Medications Section} & Medications, Medication Changes, Medication List at End of Visit, Medication, Medication Reconciliation, Preventive Medicine, Medication List, Medication List at End of Visith, Medications (active prior today), Medications (Added, Consumed or Stopped today), Medications (Added, Continued or Stopped today), Medications Changes, Medications Discontinued During This Encounter, Medications Ordered This Encounter, Medications Places This Encounter, MEDICATIONS PRESCRIBED THIS VISIT, Medications Reviewed As Of This Encounter, Meds, Outpatient Medications, Patients Medication, Preventive Medication, Previous Medications, Previous medications\\ \midrule
         \textbf{Order Info} & Orders Placed, Order Questions, Order, Order Details, Order Information, Order Providers, Order Report, Ordering Provider, Order Name, Order name, Order Number, Order Plain X-ray/Interpretation, Order Requisition, Order Tracking, Order Transmittal Tracking, Order User/Provider Detail, Order-Level Documents, Ordering Provider Information, Orders, Orders Placed This Encounter, Orders Requiring a Screening Form\\ 
\end{tabular}
    \caption{The list of sections and subsections that are normalized into one section name.}
    \label{tab:sample_cat}
% \end{table*}
% \begin{table*}[h]
%     \centering
%     \small

\end{table*}
%Electronic Health Records (EHR) offer information about patients to practitioners and enable them with easy and efficient access. Such information consists of the history of a present illness,
%physical exam findings, and progress notes. EHR is a growing volume in resources in healthcare and the challenge is these records have no standard structure and are considered unstructured and noisy data. In addition, the long length and redundancy in these documents \cite{rule2021length} make it challenging to process and structure them. Among different approaches, 
In this section, we describe the dataset and the annotation design in our study. As we described before we decided to choose section identification (SI), a method to identify sections and sub-sections in EHR documents to split them into smaller text chunks and create some structure in these unstructured data. We designed a manual annotation task to identify these sections and create categorical section types. Below we explain the annotation task design, the result, and the challenges.

\subsection{Annotation Design}

We randomly selected 100 records from a pool of one million records we have in our corpus. These records are in two forms, PDF or fax images which doctors submit to insurance companies, and hence, can arrive from any arbitrary format. We refer to these records as documents in the span of this manuscript. These documents have no standard structures and sometimes they contain multiple patients information at the same time. Six annotators with higher education and non-native speakers of English carry the annotation task. Each annotates an equal amount and random selection of these documents.

We used Label Studio\footnote{https://labelstud.io/}, an open source data labeling platform. PDF or image file of each record is uploaded to label studio and the task was to mark the section and sub-section in each file and manually enter the corresponding text of these sections and sub-sections. To instruct the annotators, we provided written instructions as well as held a video discussion session and explained the task to the annotators. 
% We obtain one annotation per PDF/image file.   

\subsection{Annotation Result}

We aggregate the sections per document to form the final section and sub-section list. A total of 912 sections and subsections are identified which makes 14 sections and sub-sections on average per document. Then one annotator, different from the ones who have annotated the documents, categorized these sections and sub-sections into more general categories based on the Consolidated Clinical Document Architecture (C-CDA) implementation guide\footnote{C-CDA contains a library of CDA templates, incorporating and harmonizing previous efforts from Health Level Seven (HL7), Integrating the Healthcare Enterprise (IHE), and Health Information Technology Standards Panel (HITSP). https://www.hl7.org/ccdasearch/}. In other words, the diverse categories are mapped to a category to unify them. This allows us to calculate IAA and be able to use the text semantic similarity method to find these sections in the unannotated documents.  A total of 464 categories are coded of which 394 of these categories have a frequency of 1 and 70 categories have a frequency of 2 or more. We provide a small sample of the most frequent categories in Table \ref{tab:sample_fre_cat} and Figure \ref{fig:section}. %\footnote{Upon acceptance of this paper the full list will be released.}. 

24 documents have been randomly selected and on each of these documents, a second annotator annotated the document. Further, we calculated the Jaccard similarity to report Inter-Annotator Agreement (IAA), The Jaccard similarity is a measure of the similarity between two sets of data. We obtained a Jaccard distance of 0.40, which is a fair agreement and an indication that the annotation task is challenging. The most diverse section and sub-section lists that each normalized into one section name are shown in table \ref{tab:sample_cat}. Notably, the diversity of these two general categories indicates the challenge involved in structuring and identifying these sections in these documents. In some cases, categories such as \textit{Order Report} or \textit{Medication Reconciliation} can be both a section and sub-section according to the annotation results. This characteristic does not enforce the decision to select the general category for these types.

% Basic
\begin{figure*}
\fbox{\begin{minipage}{40em}
You are a clinician and you read the given clinical document and identify section headers from them.
\hfill\break
Find section headers only from the clinical text.
\hfill\break
For each section header, return the answer as a JSON object by filling in the following dictionary. 
\hfill\break
\{section\_title: \/\/ string representing the section header\}
\hfill\break
Here are some clinical notes of a patient from a doctor. \#\#\# \textit{\{context\_text\}} \#\#\#
\hfill\break
\end{minipage}}
\caption{Basic Prompt Template}
\label{fig:basic}
\end{figure*}

\section{Experimental Setup}
Our task here is to take as input a document and output all the section headers found in it. For our underlying use case, we carried out testing with various LLMs like GPT-4 8k \cite{openai2023gpt4}, LLaMa-2 7B \cite{touvron2023LLaMa}, and more recent Mistral 7B \cite{jiang2023mistral} prompting strategies\footnote{CoT A\ref{fig:CoT}, One Shot A\ref{fig:One Shot} and Close Ended A\ref{fig:Close Ended} prompting strategies are elaborated in appendix \ref{sec:appendix}.} (as shown in figure \ref{fig:basic}) and contrasted them with a baseline experiment that used keyword search, regex, MedSpacy library \cite{medspacy} and the best model reported by \citet{landes2022new}. MedSpacy is a clinical NLP toolkit built on the foundation of SpaCy, specifically designed to address the unique challenges of processing and extracting information from clinical text. This enables healthcare professionals to efficiently process and derive valuable insights from unstructured medical narratives. We did not restrict the tokens and used the entire clinical note for MedSecId. We extracted the actual section header using the header span mentioned in the MedSecId annotation and used it as the ground truth for our task. 
% Due to the cost involved in running these expensive experiments, we reported our results using 10\% of randomly selected data from the corpus, similar to the MedSecID paper.  
Because of the longer length of real-world data, we used the 32k version of GPT-4 while keeping all the hyper-parameters to default such as the temperature, frequency penalty, and presence penalty to 0 and max tokens to 1000. Lastly, in this study, we utilized a privately hosted instance of GPT-4 to ensure the prevention of any potential data leakage. Prior to initiating the experiment, we implemented a thorough anonymization procedure to protect the dataset Protected health information (PHI). This involved substituting all personal identifiers, such as names, identification numbers, and ages, with fictitious entities.

Apart from the basic prompts, we also experiment with combining them with Few-Shot \cite{brown2020language} and CoT Prompting \cite{wei_chain--thought_2022} where we ask the LLM to think step-by-step along with providing an example of the clinical note and a list of headings. We keep the prompts same across all the datasets. Lastly, the evaluation metric used here is the exact match (EM) accuracy as well as precision (P), recall (R), and F1-score calculated by comparing GPT-4's output to that of ground truth in the Inside-Outside-Beginning (IOB) scheme \cite{ramshaw1999text} as used in work by \citet{landes2022new}.
Similar GPT-4 experiments were conducted on i2b2 2010 dataset but as the context length of i2b2 was smaller, in all the experiments we use GPT-4 8K. Lastly, because of cost constraints, we chose the best-performing model on above mentioned benchmarks to be evaluated against our internal real-world dataset.
%Because our approach isn't constrained to IOB scheme \cite{ramshaw1999text} as used in work by \cite{landes2022new}, we do not need to create a confusion matrix.

\section{Results} 
\begin{table*}[ht!]
    \centering
    \small
    \begin{tabular}{lccccc} \toprule
    \textbf{Method}       & \textbf{Accuracy(\%)}   & \textbf{Precision(\%)} & \textbf{Recall(\%)}  & \textbf{F1(\%)} & \textbf{EM(\%)} \\ \midrule
    Keyword Based                   & 36.07 & 100 & 36.07 & 53.01 &  36.05 \\ 
    Regex                           & 49.24 & 100 & 30.07 & 46.24 & 50.8  \\
    MedSpacy                        & 56.63 & 100  & 38.29 & 55.38 & 62.63 \\
    GPT-4 Close Ended Prompt          & 73.23 & 100 & 73.23 & 84.55 & 73.2 \\ 
    GPT-4 Chain-of-Thought (CoT)      & 94.9 & 100 & 88.62 & 93.97 & 92.47 \\
    GPT-4 Zero Shot Prompt            & 94.41 & 100 & 87.61 & 93.40 & 92.05\\
    GPT-4 One Shot Prompt             & \textbf{96.86} &100 & \textbf{92.93} & \textbf{96.24} & \textbf{96.11}\\
   % GPT One Shot CoT Prompt         & \textbf{ 97.42} & 100 & 94.18 & 97 \\
    LLaMa-2 Close Ended Prompt        & 39.96 & 100 & 39.96 & 57.10 & 39.94 \\ 
    LLaMa-2 Zero Shot Prompt          & 52.29 & 94.61 & 32.92 & 48.82 & 62.25\\ 
    LLaMa-2 One Shot Prompt           & 13.95 & 94.57 & 6.86 & 12.80 & 16.86\\
    LLaMa-2 Chain-of-Thought (CoT)    & 38.21 & 93.95 & 21.11 & 34.48 & 46.95\\
    Mistral Close Ended Prompt      & 5.24  & 100   & 5.24 & 9.96 & 5.24  \\ 
    Mistral Zero Shot Prompt        & 11.51 & 97.43 & 5.23 & 9.93 & 14.45\\ 
    Mistral One Shot Prompt         & 8.41 & 98.61 & 4.07 & 7.82 & 10.48 \\
    Mistral Chain-of-Thought (CoT)  & 11.99 & 98.61 & 5.64 & 10.67 & 15.53 \\
    BiLSTM-CRF \cite{landes2022new} & 82.2 & \textbf{95} &  95 &  95 & - \\ 
    \end{tabular}
    \caption{Results on MedSecId Corpus}
    \label{tab:res-1}
\end{table*}

\begin{table*}[ht!]
    \centering
    \small
    % \medskip
    \begin{tabular}{lccccc} \toprule
    \textbf{Method}       & \textbf{Accuracy(\%)}   & \textbf{Precision(\%)} & \textbf{Recall(\%)}  & \textbf{F1(\%)} & \textbf{EM(\%)} \\ \midrule
    Keyword Based                   & 10.98 & 100 & 8.78 & 16.14 &  69.5 \\ 
    Regex                           & 66.26 & 100 & 48.27 & 65.11  & 56.8\\
    MedSpacy                        & 38.45 & 100 & 21.92 & 35.96  & 38.14\\
    GPT-4 Close Ended Prompt          & 11.82 & 78.24 & 8.46 & 15.27 & 73.8\\ 
    GPT-4 Chain-of-Thought (CoT)      & 86.26 & 99.85& 74.65 & 85.43 & 84.33\\
    GPT-4 Zero Shot Prompt            & 89.47 & 100& 78.46 & 87.93 & 84.58 \\
    GPT-4 One Shot Prompt             & \textbf{93.03} & \textbf{100} & \textbf{85.36} & \textbf{92.10} & \textbf{89.45} \\
    LLaMa-2 Close Ended Prompt        & 88.79 & 100 & 83.57 & 91.05 & 86.54 \\ 
    LLaMa-2 Zero Shot Prompt          & 56.2 & 100 & 36.62 & 53.61 & 58.59 \\ 
    LLaMa-2 One Shot Prompt           & 30.54 & 100 & 16.75 & 28.69 & 21.2 \\
    LLaMa-2 Chain-of-Thought (CoT)    & 40.23 & 99.83 & 22.61 & 36.87 & 50.7\\
    Mistral Close Ended Prompt        & 10.41 & 100 & 6.65 & 12.48 & 19.34 \\ 
    Mistral Zero Shot Prompt          & 35.30 & 100 & 18.98 & 31.90 & 36.17 \\ 
    Mistral One Shot Prompt           & 6.58 & 100 & 3.24 & 6.29 & 7.80  \\
    Mistral Chain-of-Thought (CoT)    & 32.13 & 99.80 & 17.03 & 29.09 & 33.66 \\
   % GPT One Shot CoT Prompt         & \textbf{ 97.42} & 100 & 94.18 & 97 \\
   Maximum Entropy \cite{tepper2012statistical} & - & 91.1& 90.8  &  91 & - \\
    \end{tabular}
    \caption{Results on i2b2 Corpus. While GPT-4 has superior performance, LLaMa-2 is not far behind.}
    \label{tab:res-2}
\end{table*}

\begin{table}[h]
    \centering
    \small
 \begin{tabular}{lccccc} \toprule
    \textbf{Method}       & \textbf{A}   & \textbf{P} & \textbf{R}  & \textbf{F1} & \textbf{EM} \\ \midrule
Regex                & \textbf{67.64} & 98.69 & \textbf{51.30} & \textbf{67.51} & \textbf{71.9}\\
MedSpacy             & 5.92 & 100 & 4.13 & 7.93 & 15.72\\  
GPT-4 ZS        & 37.53 & 100 & 24.18 & 38.95 & 37.29\\
LLaMa-2 ZS      & 13.33 & 100 & 7.81 & 14.49 & 19.75\\  
Mistral ZS    & 3.67 & 100 & 1.83 & 3.60 & 5.24\\  
\end{tabular}
    \caption{Results on Real-World Corpus. ZS stands for Zero-Shot prompting}
    \label{res:real-world}
\end{table}

Even though GPT-4 was able to perform very well on open source benchmark datasets, it was unable to reach the same level of performance on our internal corpus due to its complexity as shown in table \ref{res:real-world}. Experiments showed that GPT-4 was able to achieve an accuracy of only 37\% in contrast to that of 96\% on MedSecId corpus. LLaMa-2 and MedSpacy performed equally well, in that, former achieved higher recall than latter. This can be attributed to the global knowledge encoded in the LLMs, which is not the case with MedSpacy, while on the other hand MedSpacy would be much faster to run with less overhead.
Results in table \ref{tab:res-1} and \ref{tab:res-2} show that one-shot GPT-4 \citet{openai2023gpt4} performed the best and achieved a new state of the art on MedSecId outperforming previous models by a significant margin. This unsupervised methodology beats all the supervised models on the MedSecId corpus \cite{landes2022new}. Similarly, one-shot also had a state-of-the-art performance on i2b2 2010 dataset. On the other hand, LLaMa-2 did not perform as well as GPT-4, but nevertheless had on par performance with regex. Additionally, LLaMa-2 \citet{touvron2023LLaMa} performance on i2b2 dataset came very close to that of GPT-4 itself. This disparity in performance of LLaMa-2 as well as its variation in results across the experiments leads to inconclusive results. Lastly, Mistral \cite{jiang2023mistral} performance was sub-optimal, exhibiting only a marginal improvement than a naive keyword based approach.

\begin{table*}[th!]
    \centering
    \small
    % \medskip
    \begin{tabular}{lccc} \toprule
    \textbf{Section Categories}       & \textbf{Number of Sections in Category}   & \textbf{Frequency} & \textbf{Frequency (\%)} \\ \midrule
    Assessment \& Plan &413 &958 &60.98\\
    physical exam &66 &152 &9.67\\
    Personal Info &54 &73 &4.64\\
    Medication &19 &55 &3.50\\
    History of Present Illness &3 &44 &2.80\\
    Family History &5 &40 &2.54\\
    Allergies &4 &40 &2.54\\
    Order Info &17 &38 &2.41\\
    Clinical Info &16 &36 &2.29\\
    UNKNOWN &13 &25 &1.59\\
    Additional Info &4 &18 &1.14\\
    Appointment Date &6 &15 &0.95\\
    Progress Notes &1 &15 &0.95\\
    Results &7 &12 &0.76\\
    Mental Status &6 &10 &0.65\\
    History &3 &10 &0.64\\
    Lab Results &5 &6 &0.38\\
    Alcohol Use &2 &5 &0.31\\
    Abdomen &2 &5 &0.31\\
    Referral &3 &3 &0.19\\
    Active Medication &3 &3 &0.19\\
    References &2 &3 &0.19\\
    Miscellaneous &2 &2 &0.12\\
    All Reviewer List &2 &2 &0.12\\
    Return Visit &1 &1 &0.06\\
    \end{tabular}
    \caption{Each section name is categorised to either its top-header section or a category is selected by human to represent the topic of the section. This annotation is done manually by two annotators where one selected a course-grained categories and the other selected a fine-grained categories. The one we show in this table is the coarse-grained category list, along with the number of of sections in each category, frequency, and frequency percentage. When the annotator were not able to asses a category they mark the section as \textit{UNKNOWN}}
    \label{tab:category-names}
\end{table*}

\section{Discussion}
We performed an in-depth error analysis on the subset of records that GPT-4 was unable to predict correction. Our analysis found errors in the MedSecId dataset itself, which is one of the reasons GPT-4 did not get a 100\% performance. Error analysis reveals on the rest of 2.8\% missed sections of the GPT-4 finds that 18\% of the above stated 2.8\% belong to the ``Findings'' section label and 13\% belong to the ``Image-Type'' category. Most of the documents did not have those section headers explicitly mentioned and were hidden as part of the text. Even though the precision was 100\% in i2b2 2010 dataset, the granularity of the subsections, the presence of ambiguous language, or the lack of clear markers for section boundaries could be the contributors to the slight dip in recall of the section headers. We leave fixing the issues in the dataset and advanced prompting for future work. 

Surprisingly, we found that GPT-4 was even able to extract sub-sections that were missed in the human annotations in MedSecId. This raises the question of whether GPT-4's superior performance on these datasets can be attributed to its prior exposure to them? We found out that MedSecId is derived from MIMIC dataset which forbids being used for LLM training, therefore, it is highly unlikely it was used during model training. 

Further analysis of our internal dataset revealed that high variation in the structure of the document is the root cause of such a wide gap between benchmark and our internal datasets. The original version of our data is in the form of images and PDF files. While GPT was resilient to most OCR errors it did contribute to some misspelled sections.
We acknowledge the difference in GPT's and the gold standard's approach to section title extraction. While the gold standard highlights literal text, GPT summarizes the content, potentially providing a more concise and informative overview.
Example GPT output \textit{Patient Information and Visit Details} encompasses multiple headers like \textit{Chief Complaint, History of Present Illness}, and \textit{Patient Information}. 
% We will work on refining our prompting methods .
GPT also extracted irrelevant titles as section headers \textit{Provider Information and Signature}, \textit{Page Footer}, etc. We aim to work on addressing these issues by incorporating context awareness into the title-generation process.

The major challenge in performance drop on internal dataset is due to the nature of our data itself. More specifically, there is neither standard structure nor format. The situation exacerbates with the document being an out of an OCR system which introduces numerous morphological errors. 
% Despite several efforts to clean these texts, it is still challenging to completely clean these documents.
Consequently, GPT-4's responses on our dataset are more creative and semantically similar which is something an exact match evaluation is unable to measure. 
As zero-shot was performing extremely well on public corpus and the improvement with other prompting techniques gave only minor improvements, we conducted only zero shot on our internal datasets.

Apart from conducting experiments on the state of art LLMs like GPT-4 \citep{openai2023gpt4}, we also wanted to experiment with smaller open-source models that offer flexibility. We experimented with two of the best-performing models LLaMa-2 \citep{touvron2023LLaMa} and Mistral \citep{jiang2023mistral}. However, in reality, both the open source models found it hard to follow the prompts and the outputs are not consistent. The challenges were further exacerbated when the models were required to generate results in a uniform format. Sometimes, both LLaMa-2 and Mistral would just output the summarization of the text. LLaMa-2 demonstrated a significantly superior performance than Mistral on both i2b2 and MedSecID.

Further, each section name is categorised to either its top-header section or a category is selected by human to represent the topic of the section. This annotation is done manually by two annotators where one selected a course-grained category list and other selected a fine-grained one. The one we show in table \ref{tab:category-names} is the coarse-grained category list, along with the number of sections in each category, frequency, and frequency percentage. 25 categories are created by the annotator to represent the coarse-grained categories. There are some section names that both annotators are unable to assess or select a category. These sections are categorized as \textit{UNKNOWN}. If we consider that the top nodes in an ontology network, on average each node will have 26 child nodes in this ontology. 

\section{Conclusion}
In this work, we evaluated LLMs capabilities in segmenting a clinical document into individual sections. More specifically, we show that an unsupervised GPT-4 can nearly solve the Section Identification task. Even though GPT-4 has a very high accuracy on the benchmark datasets, however, its performance on a real-world dataset has a significant lag. We further analyze the reasons for such a wide gap and find that the source dataset has cleanly defined section headers which is not the case with its real-world counterpart. To show how diverse the real-world dataset is, we further derived an ontology using another set of annotators that we share with the community at large. 

To that end, we create a harder benchmark, one that is derived from real-world data generating process. Moreover, we conducted an annotation study with five annotators to create the final dataset and found high ambiguity in the identification of headers on the newly introduced benchmark. As a takeaway, we suggest that if the source dataset or EHR is clean, then there is no need anymore to train specific supervised models to detect sections as an unsupervised LLM can perform that task. 

\section{Future Work}
After realizing the close-to-perfect performance and poor performance on the internal real world dataset of an unsupervised LLM in this study, we believe currently released datasets do not paint a clear picture of how the techniques proposed so far would perform in real world scenarios. Using our own internal dataset, we would like to fine-tune the LLM to see whether it can improve performance in a way that is comparable to open-source. Lastly, because sharing sensitive patient data is not possible, we plan to work on de-identifying and training an LLM to generate synthetic but realistic datasets which could lead to better real world benchmarks.

\section{Limitations}
One of the self-evident limitations of our approach is the reliance on GPT-4 to perform SI task. Using GPT-4 incurs both high overhead costs and significant data leakage risks if not set up properly. Therefore, the technique itself cannot be run in an isolated environment as it depends on an external API. Another drawback common with ML systems is if tomorrow new sections emerge and GPT-4 is not updated, the if will fail to capture the new section types.

\section{Ethics}
\label{ethics}
The datasets used in the study involved sensitive patient data. Therefore, we decided not to disclose the internal data. Additionally, even for the data based on MIMIC \cite{johnson2016mimic}, we used a privately hosted instance of GPT-4 that sits in a HIPAA compliant environment. Separately, the annotators were provided fully de-identified data, and the identification of the annotators themselves was anonymized during the annotation process. We have released the taxonomy at our github\footnote{\url{https://github.com/inQbator-eviCore/LLM_section_identifiers}} and kindly request the community to report any further advancements to us via email.

\clearpage
\bibliography{custom}

\begin{thebibliography}{29}
\expandafter\ifx\csname natexlab\endcsname\relax\def\natexlab#1{#1}\fi

\bibitem[{Ashok and Lipton(2023)}]{ashok2023promptner}
Dhananjay Ashok and Zachary~C Lipton. 2023.
\newblock Promptner: Prompting for named entity recognition.
\newblock \emph{arXiv preprint arXiv:2305.15444}.

\bibitem[{Bian et~al.(2023)Bian, Zheng, Zhang, and Zhu}]{bian2023inspire}
Junyi Bian, Jiaxuan Zheng, Yuyi Zhang, and Shanfeng Zhu. 2023.
\newblock \href {http://arxiv.org/abs/2309.12278} {Inspire the large language model by external knowledge on biomedical named entity recognition}.

\bibitem[{Brown et~al.(2020)Brown, Mann, Ryder, Subbiah, Kaplan, Dhariwal, Neelakantan, Shyam, Sastry, Askell et~al.}]{brown2020language}
Tom Brown, Benjamin Mann, Nick Ryder, Melanie Subbiah, Jared~D Kaplan, Prafulla Dhariwal, Arvind Neelakantan, Pranav Shyam, Girish Sastry, Amanda Askell, et~al. 2020.
\newblock Language models are few-shot learners.
\newblock \emph{Advances in neural information processing systems}, 33:1877--1901.

\bibitem[{CCSI(2022)}]{faxburden}
CCSI. 2022.
\newblock Six healthcare workflows primed for cloud faxing.
\newblock https://healthitsecurity.com/news/six-healthcare-workflows-primed-for-cloud-faxing.
\newblock Accessed: 2023-12-15.

\bibitem[{Congress(2009)}]{congress2009hr}
US~Congress. 2009.
\newblock Hr 1: American recovery and reinvestment act of 2009.
\newblock \emph{Washington, DC (February 2009)}.

\bibitem[{Devlin et~al.(2018)Devlin, Chang, Lee, and Toutanova}]{devlin2018bert}
Jacob Devlin, Ming-Wei Chang, Kenton Lee, and Kristina Toutanova. 2018.
\newblock Bert: Pre-training of deep bidirectional transformers for language understanding.
\newblock \emph{arXiv preprint arXiv:1810.04805}.

\bibitem[{Eyre et~al.(2021)Eyre, Chapman, Peterson, Shi, Alba, Jones, Box, DuVall, and Patterson}]{medspacy}
H.~Eyre, A.~B. Chapman, K.~S. Peterson, J.~Shi, P.~R. Alba, M.~M. Jones, T.~L. Box, S.~L. DuVall, and O.~V. Patterson. 2021.
\newblock {{L}aunching into clinical space with medspa{C}y: a new clinical text processing toolkit in {P}ython}.
\newblock \emph{AMIA Annu Symp Proc}, 2021:438--447.

\bibitem[{Gao et~al.(2022)Gao, Dligach, Miller, Tesch, Laffin, Churpek, and Afshar}]{gao2022hierarchical}
Yanjun Gao, Dmitriy Dligach, Timothy Miller, Samuel Tesch, Ryan Laffin, Matthew~M Churpek, and Majid Afshar. 2022.
\newblock Hierarchical annotation for building a suite of clinical natural language processing tasks: Progress note understanding.
\newblock In \emph{LREC... International Conference on Language Resources \& Evaluation:[proceedings]. International Conference on Language Resources \& Evaluation}, volume 2022, page 5484. NIH Public Access.

\bibitem[{Hersh and Hoyt(2018)}]{hersh2018health}
William~R Hersh and Robert~E Hoyt. 2018.
\newblock \emph{Health Informatics: Practical Guide Seventh Edition}.
\newblock Lulu. com.

\bibitem[{Hirohata et~al.(2008)Hirohata, Okazaki, Ananiadou, and Ishizuka}]{hirohata-etal-2008-identifying}
Kenji Hirohata, Naoaki Okazaki, Sophia Ananiadou, and Mitsuru Ishizuka. 2008.
\newblock \href {https://aclanthology.org/I08-1050} {Identifying sections in scientific abstracts using conditional random fields}.
\newblock In \emph{Proceedings of the Third International Joint Conference on Natural Language Processing: Volume-{I}}.

\bibitem[{Jiang et~al.(2023)Jiang, Sablayrolles, Mensch, Bamford, Chaplot, de~las Casas, Bressand, Lengyel, Lample, Saulnier, Lavaud, Lachaux, Stock, Scao, Lavril, Wang, Lacroix, and Sayed}]{jiang2023mistral}
Albert~Q. Jiang, Alexandre Sablayrolles, Arthur Mensch, Chris Bamford, Devendra~Singh Chaplot, Diego de~las Casas, Florian Bressand, Gianna Lengyel, Guillaume Lample, Lucile Saulnier, Lélio~Renard Lavaud, Marie-Anne Lachaux, Pierre Stock, Teven~Le Scao, Thibaut Lavril, Thomas Wang, Timothée Lacroix, and William~El Sayed. 2023.
\newblock \href {http://arxiv.org/abs/2310.06825} {Mistral 7b}.

\bibitem[{Johnson et~al.(2016)Johnson, Pollard, Shen, Lehman, Feng, Ghassemi, Moody, Szolovits, Anthony~Celi, and Mark}]{johnson2016mimic}
Alistair~EW Johnson, Tom~J Pollard, Lu~Shen, Li-wei~H Lehman, Mengling Feng, Mohammad Ghassemi, Benjamin Moody, Peter Szolovits, Leo Anthony~Celi, and Roger~G Mark. 2016.
\newblock Mimic-iii, a freely accessible critical care database.
\newblock \emph{Scientific data}, 3(1):1--9.

\bibitem[{Landes et~al.(2022)Landes, Patel, Huang, Webb, Di~Eugenio, and Caragea}]{landes2022new}
Paul Landes, Kunal Patel, Sean~S Huang, Adam Webb, Barbara Di~Eugenio, and Cornelia Caragea. 2022.
\newblock A new public corpus for clinical section identification: Medsecid.
\newblock In \emph{Proceedings of the 29th International Conference on Computational Linguistics}, pages 3709--3721.

\bibitem[{Liu et~al.(2022)Liu, Capurro, Nguyen, and Verspoor}]{LIU2022104149}
Jinghui Liu, Daniel Capurro, Anthony Nguyen, and Karin Verspoor. 2022.
\newblock \href {https://doi.org/https://doi.org/10.1016/j.jbi.2022.104149} {“note bloat” impacts deep learning-based nlp models for clinical prediction tasks}.
\newblock \emph{Journal of Biomedical Informatics}, 133:104149.

\bibitem[{Liu et~al.(2023)Liu, Yu, Zhang, Wu, Cao, Dai, Zhao, Liu, Shen, Li, Liu, Zhu, and Li}]{liu2023deidgpt}
Zhengliang Liu, Xiaowei Yu, Lu~Zhang, Zihao Wu, Chao Cao, Haixing Dai, Lin Zhao, Wei Liu, Dinggang Shen, Quanzheng Li, Tianming Liu, Dajiang Zhu, and Xiang Li. 2023.
\newblock \href {http://arxiv.org/abs/2303.11032} {Deid-gpt: Zero-shot medical text de-identification by gpt-4}.

\bibitem[{McKnight and Srinivasan(2003)}]{McKnightS03}
Larry McKnight and Padmini Srinivasan. 2003.
\newblock \href {https://knowledge.amia.org/amia-55142-a2003a-1.616734/t-001-1.619623/f-001-1.619624/a-089-1.619835/a-090-1.619832} {Categorization of sentence types in medical abstracts}.
\newblock In \emph{{AMIA} 2003, American Medical Informatics Association Annual Symposium, Washington, DC, USA, November 8-12, 2003}. {AMIA}.

\bibitem[{Nair et~al.(2021)Nair, Narayanan, Achan, and Soman}]{nair2021clinical}
Namrata Nair, Sankaran Narayanan, Pradeep Achan, and KP~Soman. 2021.
\newblock Clinical note section identification using transfer learning.
\newblock In \emph{Proceedings of Sixth International Congress on Information and Communication Technology: ICICT 2021, London, Volume 1}, pages 533--542. Springer.

\bibitem[{Nori et~al.(2023)Nori, King, McKinney, Carignan, and Horvitz}]{nori2023capabilities}
Harsha Nori, Nicholas King, Scott~Mayer McKinney, Dean Carignan, and Eric Horvitz. 2023.
\newblock Capabilities of gpt-4 on medical challenge problems.
\newblock \emph{arXiv preprint arXiv:2303.13375}.

\bibitem[{OpenAI(2023)}]{openai2023gpt4}
OpenAI. 2023.
\newblock \href {http://arxiv.org/abs/2303.08774} {Gpt-4 technical report}.

\bibitem[{Podder et~al.(2023)Podder, Lew, and Sassan}]{soap-notes}
Vivek Podder, Valerie Lew, and Ghassemzadeh Sassan. 2023.
\newblock \href {https://www.ncbi.nlm.nih.gov/books/NBK482263} {\emph{SOAP Notes}}.
\newblock StatPearls Publishing.

\bibitem[{Pomares-Quimbaya et~al.(2019)Pomares-Quimbaya, Kreuzthaler, and Schulz}]{PomaresQuimbaya2019CurrentAT}
Alexandra Pomares-Quimbaya, Markus Kreuzthaler, and Stefan Schulz. 2019.
\newblock \href {https://api.semanticscholar.org/CorpusID:197663913} {Current approaches to identify sections within clinical narratives from electronic health records: a systematic review}.
\newblock \emph{BMC Medical Research Methodology}, 19.

\bibitem[{Ramshaw and Marcus(1999)}]{ramshaw1999text}
Lance~A Ramshaw and Mitchell~P Marcus. 1999.
\newblock Text chunking using transformation-based learning.
\newblock In \emph{Natural language processing using very large corpora}, pages 157--176. Springer.

\bibitem[{Rule et~al.(2021)Rule, Bedrick, Chiang, and Hribar}]{adam_lengthy_notes2021}
Adam Rule, Steven Bedrick, Michael~F. Chiang, and Michelle~R. Hribar. 2021.
\newblock \href {https://doi.org/10.1001/jamanetworkopen.2021.15334} {{Length and Redundancy of Outpatient Progress Notes Across a Decade at an Academic Medical Center}}.
\newblock \emph{JAMA Network Open}, 4(7):e2115334--e2115334.

\bibitem[{Tepper et~al.(2012)Tepper, Capurro, Xia, Vanderwende, and Yetisgen-Yildiz}]{tepper2012statistical}
Michael Tepper, Daniel Capurro, Fei Xia, Lucy Vanderwende, and Meliha Yetisgen-Yildiz. 2012.
\newblock Statistical section segmentation in free-text clinical records.
\newblock In \emph{Lrec}, pages 2001--2008.

\bibitem[{Touvron et~al.(2023)Touvron, Martin, Stone, Albert, Almahairi, Babaei, Bashlykov, Batra, Bhargava, Bhosale, Bikel, Blecher, Ferrer, Chen, Cucurull, Esiobu, Fernandes, Fu, Fu, Fuller, Gao, Goswami, Goyal, Hartshorn, Hosseini, Hou, Inan, Kardas, Kerkez, Khabsa, Kloumann, Korenev, Koura, Lachaux, Lavril, Lee, Liskovich, Lu, Mao, Martinet, Mihaylov, Mishra, Molybog, Nie, Poulton, Reizenstein, Rungta, Saladi, Schelten, Silva, Smith, Subramanian, Tan, Tang, Taylor, Williams, Kuan, Xu, Yan, Zarov, Zhang, Fan, Kambadur, Narang, Rodriguez, Stojnic, Edunov, and Scialom}]{touvron2023LLaMa}
Hugo Touvron, Louis Martin, Kevin Stone, Peter Albert, Amjad Almahairi, Yasmine Babaei, Nikolay Bashlykov, Soumya Batra, Prajjwal Bhargava, Shruti Bhosale, Dan Bikel, Lukas Blecher, Cristian~Canton Ferrer, Moya Chen, Guillem Cucurull, David Esiobu, Jude Fernandes, Jeremy Fu, Wenyin Fu, Brian Fuller, Cynthia Gao, Vedanuj Goswami, Naman Goyal, Anthony Hartshorn, Saghar Hosseini, Rui Hou, Hakan Inan, Marcin Kardas, Viktor Kerkez, Madian Khabsa, Isabel Kloumann, Artem Korenev, Punit~Singh Koura, Marie-Anne Lachaux, Thibaut Lavril, Jenya Lee, Diana Liskovich, Yinghai Lu, Yuning Mao, Xavier Martinet, Todor Mihaylov, Pushkar Mishra, Igor Molybog, Yixin Nie, Andrew Poulton, Jeremy Reizenstein, Rashi Rungta, Kalyan Saladi, Alan Schelten, Ruan Silva, Eric~Michael Smith, Ranjan Subramanian, Xiaoqing~Ellen Tan, Binh Tang, Ross Taylor, Adina Williams, Jian~Xiang Kuan, Puxin Xu, Zheng Yan, Iliyan Zarov, Yuchen Zhang, Angela Fan, Melanie Kambadur, Sharan Narang, Aurelien Rodriguez, Robert Stojnic, Sergey Edunov, and Thomas
  Scialom. 2023.
\newblock \href {http://arxiv.org/abs/2307.09288} {Llama 2: Open foundation and fine-tuned chat models}.

\bibitem[{Uzuner et~al.(2011)Uzuner, South, Shen, and DuVall}]{uzuner20112010}
{\"O}zlem Uzuner, Brett~R South, Shuying Shen, and Scott~L DuVall. 2011.
\newblock 2010 i2b2/va challenge on concepts, assertions, and relations in clinical text.
\newblock \emph{Journal of the American Medical Informatics Association}, 18(5):552--556.

\bibitem[{Wang et~al.(2023)Wang, Sun, Li, Ouyang, Wu, Zhang, Li, and Wang}]{wang2023gpt}
Shuhe Wang, Xiaofei Sun, Xiaoya Li, Rongbin Ouyang, Fei Wu, Tianwei Zhang, Jiwei Li, and Guoyin Wang. 2023.
\newblock Gpt-ner: Named entity recognition via large language models.
\newblock \emph{arXiv preprint arXiv:2304.10428}.

\bibitem[{Wei et~al.(2022)Wei, Wang, Schuurmans, Bosma, Ichter, Xia, Chi, Le, and Zhou}]{wei_chain--thought_2022}
Jason Wei, Xuezhi Wang, Dale Schuurmans, Maarten Bosma, Brian Ichter, Fei Xia, Ed~Chi, Quoc Le, and Denny Zhou. 2022.
\newblock \href {https://arxiv.org/abs/2201.11903v6} {Chain-of-{Thought} {Prompting} {Elicits} {Reasoning} in {Large} {Language} {Models}}.

\bibitem[{Zhou et~al.(2023)Zhou, Yetisgen, Afshar, Gao, Savova, and Miller}]{zhou2023improving}
Weipeng Zhou, Meliha Yetisgen, Majid Afshar, Yanjun Gao, Guergana Savova, and Timothy~A Miller. 2023.
\newblock Improving model transferability for clinical note section classification models using continued pretraining.
\newblock \emph{medRxiv}.

\end{thebibliography}

\appendix
\section{Appendix}
\label{sec:appendix}
Figure \ref{fig:One Shot} illustrates an example of "One Shot" prompt method. It contains the segmentations and the seed list of heading found in MedSecId. In the end we present the entire patient notes received from the doctors. Figure \ref{fig:CoT} shows an example of "CoT" prompt. We observe that in this method the prompt should instruct the LLMs to think rationally and ask them to extract the section headers from the patient notes. Lastly, figure \ref{fig:Close Ended} shows an example of "Close Ended" prompt method. This method restricts the responses to be one of the 50 class labels that is obtained from MedSecId annotation.

Table \ref{tab:sec-names} demonstrates the top 50 populated section names that we observed in our corpus. The numbers are extracted from the aggregated annotation results. We observe that "Allergies", "Family History", and "Social History" are top 3 populated sections in the corpus. The full list is published in our GitHub which is provided in section \ref{ethics}. 

Figure \ref{fig:section} shows the sections categories. The annotation is done by two annotators. One annotator chooses course-grained categories and the other chooses more fine-grained categories. These categories are selected based on observation of top-header sections in the corpus and human judgment to associate these section names to their topic or category of representations. Our findings show that "Assessment \& Plan" is the most populated category with 958 sections and "Return Visit" us the least populated one with only 1 section. The sections are extracted from the aggregated annotation result of our study. Statistics such as number of sections per category, frequency, and frequency percentage is shown in Table \ref{tab:category-names}.

\begin{table*}[th!]
    \centering
    \small
    % \medskip
    \begin{tabular}{lccc} \toprule
    \textbf{Section Names}       & \textbf{Frequency}   & \textbf{Percentage (\%)}  \\ \midrule
    Allergies&	36	&2.3\%\\
Family History	&36	&2.3\%\\
Social History	&34	&2.2\%\\
Past Medical History	&29	&1.9\%\\
Physical Exam	&28	&1.8\%\\
Subjective	&25	&1.6\%\\
Objective	&24	&1.5\%\\
Plan	&24	&1.5\%\\
Surgical History	&24	&1.5\%\\
HPI	&23	&1.5\%\\
Assessment	&21	&1.3\%\\
Chief Complaint	&20	&1.3\%\\
History of Present Illness	&20	&1.3\%\\
Review of Systems	&19	&1.2\%\\
Impression	&17	&1.1\%\\
Medications	&16	&1.0\%\\
Vital signs	&16	&1.0\%\\
Additional Documentation	&15	&1.0\%\\
Progress Notes	&15	&1.0\%\\
ROS	&14	&0.9\%\\
Medication Changes	&13	&0.8\%\\
Orders Placed	&13	&0.8\%\\
Visit Diagnoses	&13	&0.8\%\\
Assessment/Plan	&12	&0.8\%\\
Current Medications	&11	&0.7\%\\
Past Surgical History	&11	&0.7\%\\
Vitals	&11	&0.7\%\\
Assessments	&10	&0.6\%\\
Examination	&10	&0.6\%\\
Musculoskeletal	&10	&0.6\%\\
Problems	&10	&0.6\%\\
Technique	&10	&0.6\%\\
Communications	&9	&0.6\%\\
Comparison	&9	&0.6\%\\
Exam	&9	&0.6\%\\
Findings	&9	&0.6\%\\
Reason for Appointment	&9	&0.6\%\\
Diagnosis	&8	&0.5\%\\
Medical History	&8	&0.5\%\\
Medication List at End of Visit	&8	&0.5\%\\
Screening	&8	&0.5\%\\
Skin	&8	&0.5\%\\
Cardiovascular	&7	&0.4\%\\
General	&7	&0.4\%\\
History	&7	&0.4\%\\
Tobacco Use	&7	&0.4\%\\
Treatment	&7	&0.4\%\\
Eyes	&6	&0.4\%\\
Instructions	&6	&0.4\%\\
Patient Information	&6	&0.4\%\\
    \end{tabular}
    \caption{Top 50 Sections Names quantified by their frequencies and percentages in the entire corpus. We observe that "Allergies", "Family History", and "Social History" are top 3 most populated sections in the corpus.}
    \label{tab:sec-names}
\end{table*}

% one-shot
\begin{figure*}
\fbox{\begin{minipage}{40em}
You are a clinician and you read the given clinical document and identify section headers from them.
\hfill\break
Find section headers only from the clinical text. 
\hfill\break
Example clinical text: \{sample\_text\}
\hfill\break
Answer \: \{ List of section headers from the corpus. \}
\hfill\break
For each section header return the answer as a JSON object by filling in the following dictionary. 
\hfill\break
\{section\_title: \/\/ string representing the section header\}
\hfill\break
Here are some clinical notes of a patient from a doctor. \#\#\# \textit{\{context\_text\}} \#\#\#
\hfill\break
\end{minipage}}
\caption{One Shot Prompt: provide examples of segmentation as well as provide a seed list of headings found in MedSecId. }
\label{fig:One Shot}
\end{figure*}

% CoT
\begin{figure*}
\fbox{\begin{minipage}{40em}
You are a clinician and you read the given clinical document and identify section headers from them.
\hfill\break
Find section headers only from the clinical text.
\hfill\break
For each section header, return the answer as a JSON object by filling in the following dictionary. 
\hfill\break
\{section\_title: \/\/ string representing the section header
\hfill\break
   CoT: \/\/ string describing thinking step by step \}
\hfill\break
Here are some clinical notes of a patient from a doctor. \#\#\# \textit{\{context\_text\}} \#\#\#
\hfill\break
\end{minipage}}
\caption{CoT Prompt: make the LLM think rationally and try to extract all possible section headers in the clinical notes}
\label{fig:CoT}
\end{figure*}

% Close ended
\begin{figure*}
\fbox{\begin{minipage}{40em}
You are a clinician and you read the given clinical document and identify section headers from them.
\hfill\break
Classify the section headers into one of the following section type labels. 
\hfill\break
section types: \{List of section types from the MedSecId training corpus.\} 
\hfill\break
If the section headers do not belong to any of the above section type labels, classify them as \'None\'.
\newline
Only print the section types identified in a list.
Here are some clinical notes of a patient from a doctor. \#\#\# \textit{\{context\_text\}} \#\#\#
\hfill\break
\end{minipage}}

\caption{Close Ended Prompt: restrict the responses to one of the 50 class labels obtained from the MedSecId annotation.}
\label{fig:Close Ended}
\end{figure*}

\end{document}